\documentclass[letterpaper, 10 pt, conference]{ieeeconf}  

\IEEEoverridecommandlockouts                              

\usepackage{amssymb}  

\usepackage{amsmath}
\usepackage{booktabs}
\usepackage{algorithm}
\usepackage{algorithmic}
\usepackage{xcolor}
\usepackage{multicol}
\usepackage{graphicx,subcaption}
\newtheorem{theorem}{\bf{Theorem}}

\title{\LARGE \bf
V-RVO: Decentralized Multi-Agent Collision Avoidance using Voronoi Diagrams and Reciprocal Velocity Obstacles
}

\author{Senthil Hariharan Arul$^{1}$ and Dinesh Manocha$^{2}$
\thanks{*This work was not supported by any organization}
\thanks{$^{1}$Senthil Hariharan Arul is with the Department of Electrical and Computer Engineering, University of Maryland, College Park, MD 20740, USA
        {\tt\small sarul1@umd.edu}}%
\thanks{$^{2}$Dinesh Manocha is with the Department of Computer Science, University of Maryland, College Park, MD 20740, USA
        {\tt\small dmanocha@umd.edu}}%
}

\begin{document}

\maketitle
\thispagestyle{empty}
\pagestyle{empty}

\begin{abstract}

We present a decentralized collision avoidance method for dense environments that is based on buffered Voronoi cells (BVC) and reciprocal velocity obstacles (RVO). Our approach is designed for scenarios with large number of close proximity agents  and provides passive-friendly collision avoidance guarantees. The Voronoi cells are superimposed with RVO cones to compute a suitable direction for each agent and we use that direction for computing a local collision-free path. Our approach can satisfy {\color{black}double-integrator} dynamics constraints and we use the properties of the BVC to formulate a simple, decentralized deadlock resolution strategy. We demonstrate the benefits of V-RVO in complex scenarios with tens of agents in close proximity. In practice, V-RVO's performance is comparable to prior velocity-obstacle methods and the collision avoidance behavior is significantly less conservative than ORCA.

\end{abstract}


\section{Introduction}
Recent advancements in multi-agent robotics support its use in last-mile delivery, warehouse inventory management, and urban surveillance. Many of these applications use  a large number of robots (e.g., a few hundred) in a decentralized manner~\cite{lifelong}.  A key challenge in these scenarios is multi-agent navigation, which includes the computation of safe, collision-free paths between agents in a shared environment. 

Prior multi-agent methods compute collision-free trajectories using centralized or decentralized methods. Centralized algorithms such as~\cite{tang2018complete,usc,dandrea} can generate collision-free trajectories simultaneously for all agents by considering them as one composite system. These techniques can provide rigorous guarantees in terms of (probabilistic) completeness or deadlock avoidance. In the worst case, their computation time can increase exponentially with the number of agents~\cite{solovey2016finding,goldenberg2014enhanced}. In practice, these methods are used for a few agents. In contrast, in a decentralized algorithm, each agent makes an independent decision to avoid an impending collision~\cite{RVO,ORCA,zhou2017fast,davis2019nh}. 
As a result, these algorithms are scalable and can handle a large number of agents. 
However, guaranteeing collision avoidance for decentralized methods is non-trivial, especially for agents with higher-order dynamics constraints. 


\subsection{Prior Work}
There is extensive work on developing decentralized collision-avoidance methods for multi-agent simulation, where each agent computes collision-free paths independently using local sensing of state. 
Our approach is based on the concept of velocity obstacles.
Velocity Obstacle (VO)~\cite{VO} computes a set of velocities that could result in a collision between agents or with dynamic obstacles. RVO~\cite{RVO} extends the VO concept by assuming that the circular agents share equal responsibility for collision avoidance by computing velocity cones for each pair of nearby agents. In the ORCA algorithm~\cite{ORCA}, the RVO constraints are linearized to reduce the feasible velocities to a convex set (i.e. ORCA constraints), and linear programming is used to quickly find a feasible solution.  ORCA provides a sufficient condition for collision-free navigation for single integrator dynamics. The simplicity of the ORCA formulation makes it possible to extend that algorithm to elliptical agents~\cite{best2016real}, double integrators~\cite{AVO},  linear agents~\cite{LQG,LQR}, non-holonomic agents~\cite{NH-ORCA}, combining with statistical inference techniques~\cite{kim2015brvo}, etc.  
However, the linear constraints in the velocity space tend to be overly conservative, especially when many agents are in close proximity or in dense scenarios with a large number of agents. As a result, there may be no feasible solution to the ORCA constraints and the resulting algorithms are unable to compute collision-free trajectories. As we take into account additional constraints corresponding to high-order dynamics or non-holonomic agents, the size of the feasible solution set decreases further and makes the approach even more conservative.
Other methods have been proposed to accelerate the performance in crowded or challenging environments using interpolating bridges in the narrow passages~\cite{he2017efficient}. In this paper, we present a new algorithm, V-RVO, that is less conservative than ORCA and has similar runtime performance.

Buffered Voronoi cell (BVC)~\cite{zhou2017fast} is an efficient decentralized method, which can compute collision-free trajectories for single integrator agents. In BVC, collision avoidance is performed by reasoning in the position space using only the agent's position information. Robots deviate from the goal direction only when the agents are in close proximity. This is in contrast to velocity obstacle methods, which identify that the current velocity would result in future collision and perform an avoidance maneuver. BVC's guarantees need not translate to high-order dynamics because the agent's bounded acceleration could regard a computed path as infeasible.  An MPC planner that uses BVC and the braking distance for navigation in noisy scenarios has been proposed~\cite{zhuBVC}. This is similar to our method where we use BVC and braking distance for {\em{passive guarantees}}. A decentralized multi-agent rapidly-exploring random tree (DMA-RRT) is presented in~\cite{desaraju2011decentralized}, where an RRT path planner and a token passing method are used for replanning. 


Other techniques for collision avoidance between multiple agents are based on a time-to-collision model~\cite{davis2019nh,forootaninia2017uncertainty}. Another guaranteed technique is based on inevitable collision states (ICS)~\cite{ICS} and computes a set of agent states that have no collision-free trajectories for an infinite time horizon. 
ICS provides a theoretical guarantee in terms of collision avoidance, but this method can be very conservative and could regard  the entire workspace as forbidden, even in the presence of a few agents~\cite{ICSfov}. 
\cite{macek2009} considers a relaxed safety condition  known as passive motion safety, where agents cannot collide if they have a non-zero velocity. 


\subsection{Main Contributions}
We present a novel multi-agent collision avoidance algorithm (V-RVO) that combines the benefits of velocity-space methods (e.g., velocity obstacles) and position-based methods (e.g., BVC). 
We use the velocity obstacle constraints for each agent to avoid a collision by reasoning over a time-horizon.
At the same time, we exploit the characteristics of BVC to guarantee collision avoidance for higher-order dynamics. 


In our method, each agent computes a buffered Voronoi cell (BVC) based on the neighbor's position information. We superimpose the RVO cones constraints on the BVC to compute an appropriate goal point on the BVC boundary. Next, we compute a \textit{shrunken} BVC by considering the agent's (and the moving obstacles') stopping distance. We refer to this retracted Voronoi Cell as the {\em braking-aware buffered} Voronoi cell (baBVC). We compute the braking distance and control input by  using the kinematic equations for each agent. Our V-RVO method computes a safe path for each agent inside the baBVC.
We prove that baBVC has the property that any agent on the edges of baBVC can incorporate a braking maneuver to halt before reaching the BVC edges.  We leverage the property that an agent's BVC is disjoint from its neighbors to provide the collision avoidance guarantees for agents with single and double integrator dynamics. Furthermore, we propose a simple deadlock resolution strategy.
As compared to prior methods, V-RVO has the following benefits:
\begin{itemize}
    \item V-RVO retains many benefits of ORCA, like handling high-order dynamics and anticipatory collision avoidance. The use of baBVC makes it possible to compute collision-free trajectories in dense scenarios, where ORCA is either overly conservative or fails (see Figs.~\ref{fig:70}).
    \item Unlike ORCA and BVC, V-RVO can provide safety guarantees for second order agents (Figure~\ref{fig:DI}). 
\end{itemize}

We have implemented our method and compared its performance with ORCA on agents with single- and second-order dynamics for dense scenarios. The average running time for V-RVO  is a few milliseconds per agent and is about $3$X slower than ORCA.  {\color{black} In practice, V-RVO can compute collision-free trajectories for many challenging benchmarks with $25-70$ agents, where ORCA tends to fail.} 
 

\section{Background and Problem Formulation}
In this section, we give an overview of various concepts used in our approach. The symbols and notations used in the paper are summarized in Table 1.
\begin{table}
\centering
\caption{Symbols and notation used in the paper}
\begin{tabular}{lrr}
\toprule
Notation  & Definition \\
\midrule
$A_i$ & Refers to the $i^{th}$ agent\\
 $\mathbf{p}_{i}$ & 2-D position of $A_i$ $[p_{i,x}, p_{i,y}]$\\
 $\mathbf{g}_i$ & 2-D goal position of $A_i$\\
 $\mathbf{v}_i$ & 2-D velocity of $A_i$ $[v_{i,x}, v_{i,y}]$\\
$\mathbf{u}_i$ & Control Input of the $A_i$ $[a_{i,x}, a_{i,y}]$ \\
${R}_i$ & Radius of $A_i$'s enclosing circle \\
$\mathcal{N}_i$ & Neighbors of $A_i$\\
$\bar {\mathcal{N}_i}$ & Neighbors sharing a Voronoi edge with $A_i$\\
$\mathcal {V}_i, \bar{\mathcal {V}}_i$ & Voronoi and Buffered Voronoi cell for $A_i$ \\ 
$\partial\bar{\mathcal {V}}_i$ & Boundary of the set $\bar{\mathcal {V}}_i$ \\ 
$\epsilon_p, \epsilon_v$ & Small positive constants\\
\bottomrule
\end{tabular}
\vspace*{-0.13in}
\label{tab:booktabs}
\end{table}
\subsection{Problem Formulation}
We consider an environment with $N$ agents moving in a shared workspace $\mathcal{W} \subset \mathcal{R}^2$. Each agent $A_i$, where $i\in\{1, 2, ... N\}$, has its bounding geometric shape approximated as a circle of radius $R_i$. Two agents $A_i$ and $A_j$ are collision-free if
\begin{equation*}
    \Vert \mathbf{p}_i - \mathbf{p}_j \Vert_2 \geq R_i + R_j.
\end{equation*}
That is, the agents are collision-free if the inter-agent distance between $A_i$ and $A_j$ is greater than the sum of their radii. Our goal is to compute a collision-free path for each agent towards its goal position in a decentralized manner.
Furthermore, our goal is to provide passive-friendly collision avoidance guarantees for agents with higher-order dynamics.

\subsection{Assumptions}
We assume each agent has the exact state information, including positions and velocities, of the neighboring agents. The environment may consist of  other static and dynamic obstacles. The exact position of the dynamic obstacle is known, though no assumptions are made about its trajectories. We also assume that the maximum velocity and acceleration of such agents are known a priori.

\begin{figure}[thbp]
    \centering
    \includegraphics[width=0.23\textwidth, trim={1.5cm 1.5cm 1.5cm 1.0cm},clip]{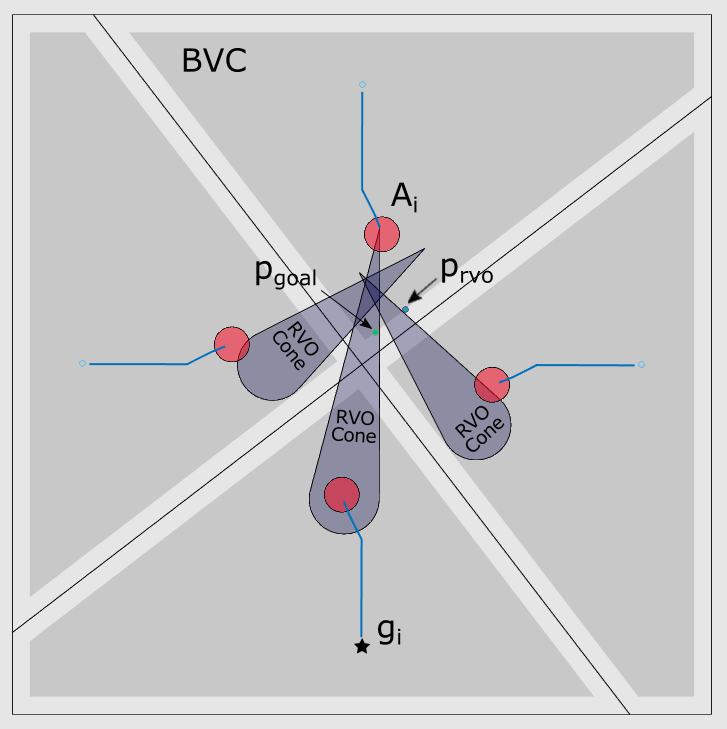}
    \vspace*{-0.09in}
    \caption{ The figure illustrates a navigation scenario with four agents moving to their antipodal positions. The three RVO cones, corresponding to three neighbors of $A_i$, are denoted by the \textit{blue} conical regions and are superimposed onto the \textit{gray} BVC region. The $\mathbf{p}_{goal}$ (green point) on the BVC boundary is in the direction of the goal ($\mathbf{g}_i$). Since this point lies in the RVO cone, we compute the closest point on the BVC boundary outside the RVO cones. Point $\mathbf{p}_{rvo}$ is denoted in \textit{blue} and the agent is directed towards this point to avoid a collision. }
     \vspace*{-0.09in}
    \label{fig:method}
\end{figure}

\subsection{Passive Safety Guarantees}
\textit{Passive safety} is a relaxed safety condition where the agent cannot collide while it is in motion~\cite{macek2009}. A collision can occur only in situations where the agent is at rest and a moving obstacle collides with the agent. In \textit{passive-friendly safety}, in addition to \textit{passive safety}, the agent provides sufficient time for the moving obstacle to stop or replan to avoid collisions. However, a collision with a moving obstacle is possible if they plan on colliding. 
Passive-friendly safety also helps to operate safely in the presence of heterogeneous agents that do not use V-RVO.

\subsection{Buffered Voronoi Cell (BVC)}\label{BVC}
We use the notion of computing Buffered Voronoi Cell (BVC),  proposed by~\cite{zhou2017fast}. BVC is a contracted Voronoi region 
such that an agent whose center point is on the edge of a BVC has its bounding circle inside the Voronoi cell. Given $\mathcal{N}$ agents on a 2-D plane, the buffered Voronoi cell corresponding to an agent $A_i$ is given as
\begin{small}
\begin{equation} 
\bar{\mathcal {V}}_i= \bigg\lbrace {\bf p}\in \mathcal {R}^2| \bigg ({\bf p}-\frac{{\bf p}_i+{\bf p}_j}{2}\bigg)^\text{T}{\bf p}_{ij} + R_\text{i}\Vert {\bf p}_{ij}\Vert \leq 0,\forall j\ne i\bigg\rbrace. 
\end{equation}
\end{small}
Here, $\bf{p}_{ij} = \bf{p}_i - \bf{p}_j$. The method plans a collision-free path for the agent $A_i$ that is constrained to lie within $\bar{\mathcal{V}}_i$. 

\subsection{Reciprocal Velocity Obstacle}\label{RVO}
Reciprocal Velocity Obstacle (RVO) computes a set of relative velocities between two agents that can result in a future collision. Let us consider two agents $A_i$ and $A_j$ in a shared workspace $\mathcal{W}$. The RVO can be geometrically defined as
\begin{align}
\begin{split}
    RVO_j^i(v_j,v_i) = \{ \mathbf{v} | \exists t \in [0,\tau]
     :: t\mathbf{v} \in {\bf{D}} (\mathbf{p}_{ji}, R_{ij} )\}.
\end{split}
\end{align}
Here, $    {\bf{D}} (\mathbf{P},r) = \{ \mathbf{q} \text{ } | \text{ } \Vert \mathbf{q} - \mathbf{p} \Vert \le r \}$ represents a disk of radius $r$ and center $\mathbf{p}$. The variable $\mathbf{p}_{ji}$ represents the relative position given by $ \mathbf{p}_{j} - \mathbf{p}_{i}$ and $R_{ij}$ represents the combined radius given by $R_i + R_j$.
If $A_i$ chooses a velocity outside the RVO induced by $A_j$, the agent is guaranteed to be collision-free, provided the trajectories of both agents are governed by single-integrator dynamics.

\section{V-RVO: Our Hybrid Navigation Algorithm}

In this section, we describe our multi-agent navigation method. {\color{black}This includes computing the BVC for each agent, followed by RVO superposition and baBVC computation.}
Algorithm~\ref{alg:algorithm} highlights our proposed method.

\subsection{RVO Superposition}\label{sec:rvosup}

In our method, we combine RVO constraints with BVC to compute a collision avoiding direction. 
Let us consider an agent $A_i$. At each time step, the agent constructs its BVC ($\Bar{\mathcal{V}}_i$), and for every {\color{black}$A_j \in \mathcal{N}_i$} we construct an RVO cone w.r.t. each nearby agent. Because the RVO constraints~(Equation~\ref{RVO}) are defined in the relative velocity space, each cone is transformed into the velocity space of $A_i$. The origin of the velocity space is the center of $A_i$. Thus, the RVO cones are superimposed onto the constructed BVC with $\mathbf{p}_i$ as the origin for the velocity space.

The union of the $\partial\bar{\mathcal {V}}_i$ sections outside the RVO cones provides safe directions of computing the velocity, as they are outside the RVO cones and within the BVC. We denote this union of $\partial\bar{\mathcal {V}}_i$ sections 
by $\partial \bar{\mathcal {V}}_{CF,i}$ and expressed as:
\begin{equation}
    \partial \bar{\mathcal {V}}_{CF,i} = \partial \bar{\mathcal {V}}_{i} - \bigcup_{j \in \mathcal{N}_i} \partial \bar{\mathcal {V}}_{i} \cap RVO^i_j(v_j, v_i).
\end{equation}
A point in the set $\partial \bar{\mathcal {V}}_{CF,i}$ with the least angular deviation from the goal direction is computed and is denoted by $\mathbf{p}_{rvo,i}$: 
\begin{small}
\begin{equation}
   \mathbf{p}_{rvo,i} = \operatorname*{argmin}_{\mathbf{p} \in \partial \bar{\mathcal {V}}_{CF,i}} {\arccos{\bigg( \frac{\mathbf{p}.(\mathbf{p}_i - \mathbf{g}_i)}{\Vert \mathbf{p} \Vert . \Vert \mathbf{p}_i - \mathbf{g}_i \Vert}\bigg)}}.
\end{equation}
\end{small}
The agent $A_i$ moves towards the $\mathbf{p}_{rvo,i}$ to avoid collisions with other agents. In order to provide passive-friendly guarantees, the agent's trajectory is planned such that it can brake and stop at the point $\mathbf{p}_{rvo,i}$. Figure~\ref{fig:method} illustrates a simple scenario with four agents.

\subsection{Braking-aware BVC}\label{sec:babvc}

As mentioned in Section~\ref{BVC}, the BVC is generated by buffering the Voronoi cell edges by the agent's radius. Since an agent with single integrator dynamics can instantaneously change its velocity, the agent is guaranteed to be collision-free inside the BVC. For agents with double integrator dynamics the BVC is further buffered based on the minimum braking distance of the agent and the moving obstacle. 

\begin{figure}
    \centering
    \includegraphics[width=0.2\textwidth]{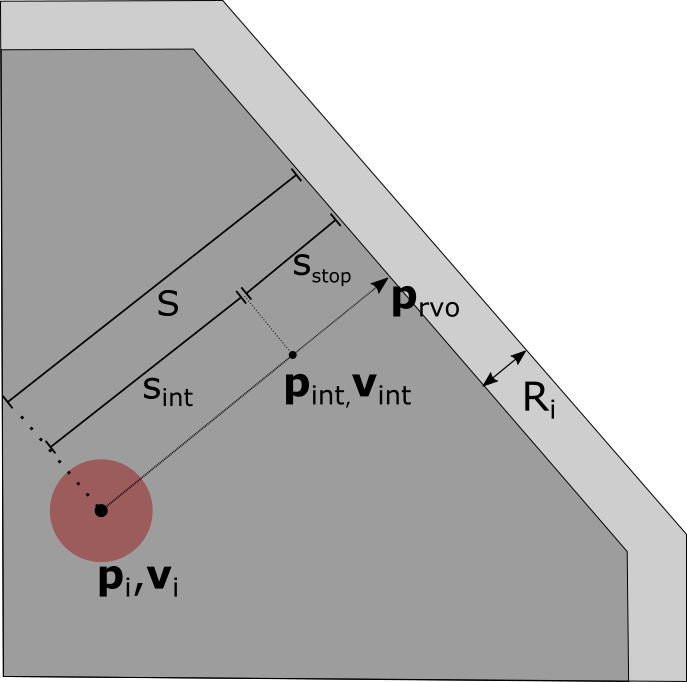}
    \caption{We show an agent (red circle), Voronoi cell (gray region) and the BVC (dark gray) computed with the agent's radius. $\mathbf{p}_{int}$ and $\mathbf{v}_{int}$ are the position and velocity at an intermediate point computed from Equation~\ref{eqn:babvc}. $s_{int}$ and $s_{stop}$ are the displacement vectors.}
    \vspace*{-0.1in}
    \label{fig:babvc}
\end{figure}

We use the kinematic equations of the agent to compute the buffering distance, as shown in Figure~\ref{fig:babvc}. Let us consider a time horizon $t_h$. The agent $A_i$ moves to a position $\mathbf{p}_{rvo,i}$ as computed using RVO constraints. Since the agent has to halt at a point on the edge of the BVC to remain within its Voronoi cell, we plan a path from $\mathbf{p}_i$ to $\mathbf{p}_{rvo,i}$ using the kinematic constraints. Consider a third position $\mathbf{p}_{int,i}$ within the BVC such that the agent at $\mathbf{p}_{int,i}$ with a velocity $\mathbf{v}_{int,i}$ can brake and stop at $\mathbf{p}_{rvo,i}$ in time $t_b$. We compute the velocity $\mathbf{v}_{int,i}$ by solving the problem separately in $X$ and $Y$ coordinates. 
\begin{align}
s_{int,x} = v_{x,i}t_h + \frac{1}{2t_h}\big( v_{int,x,i} - v_{x,i} \big) t_h^2, \label{sx}\\
s_{stop,x} = \frac{v_{int,x}^2}{2a_{max}}. \label{sstop}
\end{align}
Here, $s_{int,x}$ is the distance travelled while converting the velocity from $\mathbf{v}_{x,i}$ to $\mathbf{v}_{int,x}$, and $s_{stop,x}$ is the {\color{black}distance} travelled when the agent decelerates to a stop. Using Equations~(\ref{sx}) and~(\ref{sstop}) along with the relationship $s_x = s_{int,x} + s_{stop}$ ({\color{black}{ also there abs sum should be sx}}), we arrive at the following quadratic equation.
\begin{equation}\label{eqn:babvc}
{v}_{int,x,i}^2 + (a_{max}t_h).{v}_{int,x,i} + (u_{x,i}a_{max}t_h-2a_{max}s).
\end{equation}
The velocity $\mathbf{v}_{int,i}$ is computed from the quadratic equation by choosing the velocity with the least angular deviation from the goal direction. The position $\mathbf{p}_{int,i}$ can be computed using $\mathbf{v}_{int,i}$.

We assume low-velocity moving obstacles are present in the environment. The BVC is buffered by a distance $d_{obs}$ to provide enough stopping distance for the low-velocity moving obstacle, which is computed as: 
\begin{equation}
    d_{obs} = s_{\mathcal{O}} - \frac{\mathbf{v}_{max,obs}^2}{2a_{max}}.
\end{equation}\label{dobs}
Here, $S_\mathcal{O}$ denotes the shortest distance between the obstacle's center and $\partial\bar{\mathcal{V}_i}$.
We use this buffered BVC as the input from computing $\mathbf{p}_{rvo,i}$ in Section~\ref{sec:rvosup}.
This helps provide passive-friendly collision avoidance guarantees. This results in a \textit{contracted} BVC region, which we refer to as braking-aware buffered Voronoi cell (baBVC). We assume $S_\mathcal{O} \ge \frac{\mathbf{v}_{max,obs}^2}{2a_{max}}$ for this computation to maintain $d_{obs} \ge 0$.

\subsection{Multi-Agent Navigation}
\begin{figure*}[ht]
  \subfloat{
	\begin{minipage}[c][0.5\width]{
	   0.3\textwidth}
	   \centering
	   \includegraphics[width=0.8\textwidth, height=3.2cm, trim={2cm 1cm 3cm 1.5cm},clip]{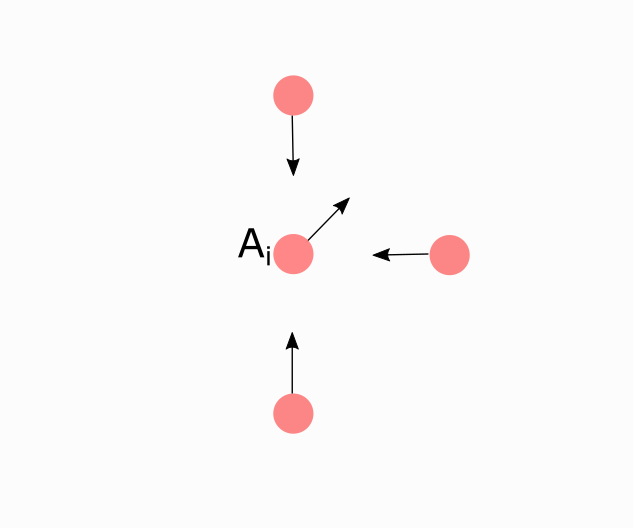}
	   \label{fig:subim1}
	\end{minipage}}
 \hfill 	
  \subfloat{
	\begin{minipage}[c][0.5\width]{
	   0.3\textwidth}
	   \centering
	   \includegraphics[width=0.8\textwidth, height=3.2cm, trim={3cm 2cm 1cm 0cm},clip]{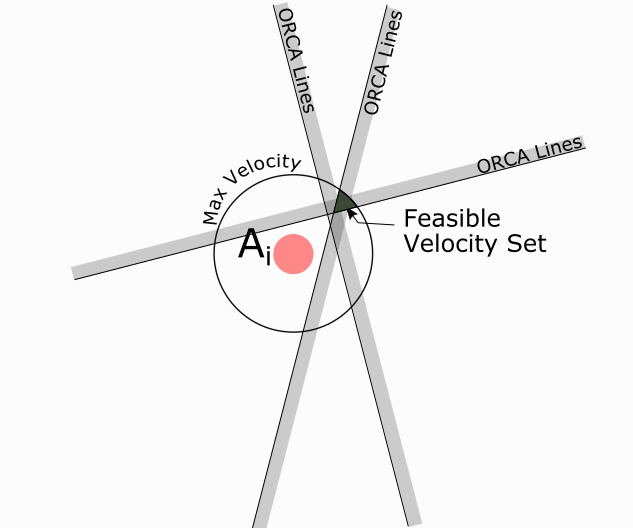}
	   \label{fig:subim2}
	\end{minipage}}
 \hfill	
  \subfloat{
	\begin{minipage}[c][0.5\width]{
	   0.3\textwidth}
	   \centering
	   \includegraphics[width=0.8\textwidth, height=3.2cm, trim={2cm 1cm 3cm 1.5cm},clip]{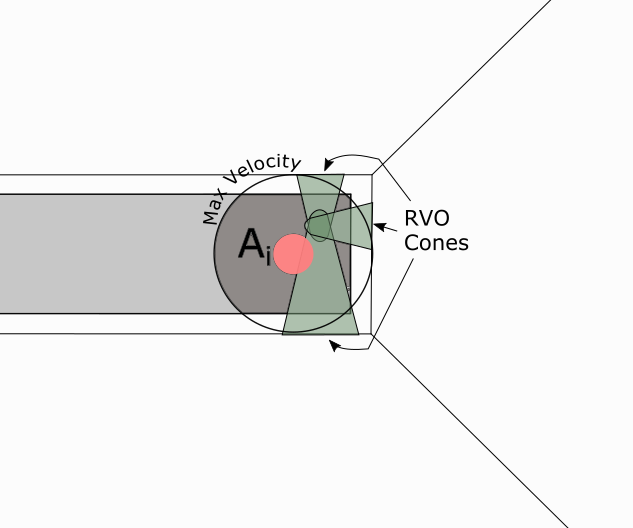}
	   \label{fig:subim3}
	\end{minipage}}
	\vspace{0.1in}
\caption{(Left) image illustrates a navigation scenario with 4 agents (red disks) and their current velocities (denoted by arrows). (Center) image illustrates the feasible velocity set for $A_i$ computed using linear ORCA constraints. In the (right) image the agent's local trajectory is computed using our V-RVO algorithm. The \textit{green} conical region are the RVO constraints superimposed onto the BVC.  The feasible regions and directions computed using V-RVO (shown in dark \textit{grey}) is larger than that compared to ORCA (center image).}
\end{figure*}

In a multi-agent scenario, all agents compute the positions $\mathbf{p}_{rvo,i}$ and $\mathbf{p}_{int,i}$ as described in Sections~\ref{sec:rvosup} and~\ref{sec:babvc}. For agents with single integrator dynamics, the positions $\mathbf{p}_{rvo,i}{} = {}\mathbf{p}_{int,i}$ as the velocity can be instantaneously changed to zero.
The control input (velocity) applied to the agent is given as
\begin{equation}
    \mathbf{u}_i = \Vert v_{max} \Vert \frac{\mathbf{p}_{rvo,i} - \mathbf{p}_{i}}{\Vert \mathbf{p}_{rvo,i} - \mathbf{p}_{i} \Vert}.
\end{equation}
If $\Vert \mathbf{p}_{rvo,i} - \mathbf{p}_{i} \Vert \le 1$, we apply $\mathbf{u}_i = \mathbf{p}_{rvo,i} - \mathbf{p}_{i}$.

For scenarios with double integrator agents, the control input ($\mathbf{u}_i$) is acceleration. A constant acceleration for the time horizon $t_h$ is computed as follows:
\begin{equation}\label{eqn:diu}
    \mathbf{u}_i = \frac{\mathbf{v}_{int,i} - \mathbf{v}_{i}}{t_h}.
\end{equation}
For double integrator agents, we also consider acceleration bounds. The path between $\mathbf{p}_{int}$ and $\mathbf{p}_i$ may not be a straight line. Thus, we verify that next position $\mathbf{p}_{t,i}'$ planned for every time step $t$ in $[0 ... t_h]$ (considering control input from Equation~\ref{eqn:diu}) lies within $\Bar {\mathcal{V}}_i$. {\color{black} $t_h$ can be varied to account for the acceleration bound.}

\begin{figure}
    \centering
   \includegraphics[width=.24\textwidth]{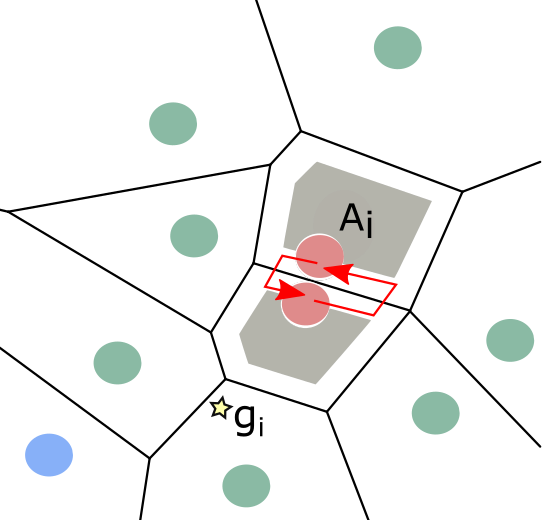}
   \vspace*{-0.09in}
    \caption{A Dense scenario with 10 agents (circles). The agent $A_i$ is deadlocked and unable to reach its goal location $g_i$ as its neighbors have reached their goal position. The neighbors in $\texttt{HOLD}$ mode are shown in $\textit{green}$ color. $A_i$ chooses a neighbor and switches its location following the \textit{red} path. The \textit{gray} region corresponds to their respective BVCs. 
    }
    \label{fig:deadlockExp}
\end{figure}

\subsection{Collision-Free Guarantees}
We prove that agents trajectories computed using V-RVO can provide {\em{passive-safety}} guarantees.
\begin{theorem}
Agents navigating in multi-agent scenarios using Algorithm I are guaranteed to be collision-free. 
\end{theorem}

\textbf{Proof:} 
Let us consider $N$ agents operating in a workspace $\mathcal{W}$. Let $\mathcal{V}_i$ be the Voronoi region corresponding to agent $A_i$. BVC region ($\bar{\mathcal{V}_i}$) is constructed by retracting the Voronoi edges by a distance equal to the agent's radius. Thus,
$\Bar{\mathcal{V}_i} \subset  \mathcal{V}_i$ and BVC region between the agents is disjoint. That is, $\Bar{\mathcal{V}}_i \cap \Bar{\mathcal{V}}_j = \emptyset \quad \forall i,j\in N, j\ne i$.

Consider the agents at their initial position $\mathbf{p}_i \quad \forall i \in N$ (at $t = t_0$). Let the corresponding region occupied by the agent geometry be $\mathcal{A}(\mathbf{p}_i)$. Assuming agents are collision-free at $t = 0$, and from the definition of BVC
\begin{multline}
     \Vert \mathbf{p}_i - \mathbf{p}_j\Vert \ge R_i + R_j \quad \forall j \in N/\{i\} \\ \implies
      \mathcal{A}(\mathbf{p}_i) \subset {\mathcal {V}}_i \iff \mathbf{p}_i \in \bar{\mathcal {V}}_i.
\end{multline}

Provided $A_i$ chooses its next position $\mathbf{p}_i' \in \bar{\mathcal {V}}_i$, then $\mathcal{A}(\mathbf{p}_i') \subset {\mathcal {V}}_i$ and the agent continues to be collision free in the next time step. 

For a single integrator agent, the velocity ($\mathbf{v}_i$) can be instantaneously modified.  Hence, the path between $\mathbf{p}_i$ and $\mathbf{p}_i'$ is a straight line ($\mathbf{p}_i\mathbf{p}_i'$). The agent's path is collision-free from the convexity of $\bar{\mathcal {V}}_i$ as $$  \mathbf{p}_i,\mathbf{p}_i' \in \bar{\mathcal {V}}_i \implies \mathbf{p}_i\mathbf{p}_i' \subset \bar{\mathcal {V}}_i \implies \mathcal{A}_i(\mathbf{0}) \oplus \mathbf{p}_i\mathbf{p}_i' \subset \mathcal{V}_i .$$
Here, $\mathbf{0}$ is the origin and $\oplus$ denotes Minkowski sum.

For double integrator agents, we compute a constant acceleration from Equation~(\ref{eqn:diu}) applied for  duration $t_h$. The acceleration moves the agent to an intermediate point $\mathbf{p}_{int, i}$ such that the agent can decelerate to a stop at $\mathbf{p}_{rvo,i} \in \partial \Bar{\mathcal{V}}_i$. Computing the agent's future positions $\mathbf{p}_{t,i}'$ using~(\ref{eqn:diu}) such that $\mathbf{p}_{t_h,i}' = \mathbf{p}_{int,i}$ and assuming a straight-line movement between time-steps. The agent's path is collision-free as
\begin{multline}
\mathbf{p}_{t,i}' \in \bar{\mathcal{V}_i} \implies \mathcal{A}(\mathbf{p}_{t,i}') \subset \mathcal{V}_{i} \implies \\ \mathcal{A}_i(\mathbf{0}) \oplus \mathbf{p}_{t-1,i}'\mathbf{p}_{t,i}' \subset \mathcal{V}_i \quad \forall t \in [1 ... t_h].
\end{multline}

The above conditions when satisfied provides {\em{passive}} collision avoidance guarantees. Since all agents in the workspace $\mathcal{W}$ use V-RVO, the agents would brake to stop at a point on $\partial \Bar{\mathcal{V}}_i$. Thus, agent do not collide with each other.

In the presence of low-velocity moving obstacles, the BVC is initially buffered as mentioned in Equation~\ref{dobs}. This provides a buffered distance for the moving-obstacle to brake without colliding. In this manner, we extend the passive guarantees to the passive-friendly.

\begin{algorithm}[tb]
\footnotesize
\caption{V-RVO Navigation Algorithm}
\label{alg:algorithm}
\textbf{Input}: $\mathbf{p}_i, \mathbf{v}_i \quad i \in \mathcal{N}$\\
\textbf{Output}: $\mathbf{u}_i \quad i \in \mathcal{N}$
\begin{algorithmic}[1] 
\FOR{$i \in \mathcal N $}
\STATE Compute BVC boundary ($\partial \Bar{\mathcal{V}}_i$) 
\STATE Compute RVO cones 
\STATE $\partial \Bar{\mathcal{V}}_{CF,i} = \partial \Bar{\mathcal{V}}_i$
\FOR{$j \in \text{RVO Cones}$}  
\STATE $\partial \Bar{\mathcal{V}}_{CF,i} \gets \partial \Bar{\mathcal{V}}_{CF,i} \cap$ {BVC edge region not inside} $j^{th}$ RVO cones \COMMENT{// compute collision-free $\partial \bar{\mathcal{V}}$ section}
\ENDFOR
\STATE $ \mathbf{p}_{rvo,i} = \operatorname*{argmin}_{\mathbf{p} \in \partial \bar{\mathcal {V}}_{CF,i}} {\arccos{\bigg( \frac{\mathbf{p}.(\mathbf{p}_i - \mathbf{g}_i)}{\Vert \mathbf{p} \Vert . \Vert \mathbf{p}_i - \mathbf{g}_i \Vert}\bigg)}}$ \COMMENT{// position in $\partial \Bar{\mathcal{V}}_{CF,i}$ with least angular deviation from goal direction}
\STATE Compute $\mathbf{p}_{slow,i}, \mathbf{v}_{slow,i}$ from Equation~(\ref{eqn:babvc})
\STATE $\mathbf{u}_i = \frac{\mathbf{v}_{slow,i} - \mathbf{v}_{i}}{t_h}$ \COMMENT{// compute control input}
\ENDFOR
\STATE \textbf{return} $\mathbf{u}_i$
\end{algorithmic}
\end{algorithm}

\section{Deadlock}
In this section, we present our distributed deadlock resolution strategy based on the constructed BVC. An illustrative example is provided in Figure~\ref{fig:deadlockExp}, and the pseudo-code is summarized in Algorithm~\ref{alg:deadlock}.

\subsection{Deadlock Identification}\label{deadID}
Deadlock occurs in situations where multiple agents block each other such that one or more agents are unable to reach their goal, and instead remain stationary (with zero velocity) to avoid a collision. In our method, each agent chooses a point on the boundary of its baBVC and travels towards it. When an agent is in a deadlock, it is yet to reach its goal position, and it is at a point on $\partial \Bar{\mathcal{V}}_i$ with a zero velocity.

\subsection{Deadlock Resolution}
To resolve deadlocks, we define three operating modes for the agents. Each agent $A_i$ can be in one of the three modes: $\texttt{DEADLOCK}$, $\texttt{HOLD}$, or $\texttt{DEFAULT}$. In general, all agents are in the $\texttt{DEFAULT}$ mode, where they are either moving towards their goal, avoiding collisions, or have reached their goal position. The agents in the $\texttt{HOLD}$ mode have their velocities set to zero and thus remain at the current position until the mode is modified to $\texttt{DEFAULT}$. When an agent is identified to be in a deadlock (as mentioned in the Section~\ref{deadID}), the agent's mode is modified to $\texttt{DEADLOCK}$. 

Consider an agent $A_i$ whose mode corresponds to $\texttt{DEADLOCK}$. If the agent $A_i$ has neighboring agents with either $\texttt{DEADLOCK}$ or $\texttt{DEFAULT}$ modes with zero velocity, then it initiates a switch to potentially solve the deadlock. 
\begin{itemize}
    \item The agent initially identifies the neighboring agent ($A_k$) that is closest to the direction of the agent's goal location. 
    Since $A_i$ and $A_k$ are neighboring agents, they share a common Voronoi edge. 
    
    \item The states of the neighbors $\bar {\mathcal{N}}_i$ and $\bar{\mathcal{N}}_k$ are set to $\texttt{HOLD}$. Since the $A_i$ and the agents in $\bar{\mathcal{N}}_i$ share Voronoi edges, setting the agents in $\bar{\mathcal{N}}_i \cup \bar{\mathcal{N}}_k$ to $\texttt{HOLD}$ ensures that the union of BVCs of $A_i$ and $A_k$ could be used for the subsequent time steps. 
    \item We plan a path between the agents $A_i$ and $A_k$ across their common Voronoi edge such that the positions of $A_i$ and $A_k$ are interchanged. (Algorithm~\ref{alg:deadlock}, line~\ref{line:switch}). 
    
    \item Once this position switch is completed, the agent state is reset to $\texttt{DEFAULT}$. 
\end{itemize}

\begin{algorithm}[h]
\footnotesize
\caption{Deadlock Resolution}
\label{alg:deadlock}
\begin{algorithmic}[1] 
\IF{$\Vert \mathbf{p}_i - \mathbf{g}_i \Vert_2 \ge \epsilon_p \text{ and } \Vert \mathbf{v}_i\Vert \le \epsilon_v$}
\STATE $Mode_{i} = \texttt{DEADLOCK}$
\STATE $k = \operatorname*{argmin}_{j \in \mathcal{N}_i}{\arccos{\bigg( \frac{(\mathbf{p}_i - \mathbf{g}_i).(\mathbf{p}_i - \mathbf{p}_j)}{\Vert \mathbf{p}_i - \mathbf{g}_i \Vert . \Vert \mathbf{p}_i - \mathbf{p}_j \Vert}\bigg)}}$\label{inline}\\
\STATE $Mode_{j} = \texttt{HOLD} \quad \forall j \in \bar{\mathcal{N}}_i \cup \bar{\mathcal{N}}_{k}$
\STATE $\texttt{SwitchAgent}(i, k)$\label{line:switch}
\STATE $Mode_{j} = \texttt{DEFAULT} \quad \forall j \in \bar{\mathcal{N}}_i \cup \bar{\mathcal{N}}_{k} \cup i \cup k$
\ENDIF
\end{algorithmic}
\end{algorithm}


\section{Result}
In this section, we describe our implementation and highlight the performance of V-RVO in different scenarios. We also compare with ORCA.

\subsection{Experimental Setup}
We perform our evaluation on a $2.7$ GHz Quad-Core Intel Core $i7$ processor with $16$ GB of memory. 
The agents and obstacles in our scenarios are circular agents of radius $0.25$m.
The agents have maximum velocity and acceleration of $2m/s$ and $1m/s^2$ respectively. We use a time step of 0.1 seconds for our simulation results.

\subsection{Simulation Results}
\begin{figure}[h]
  \includegraphics[width=.32\columnwidth]{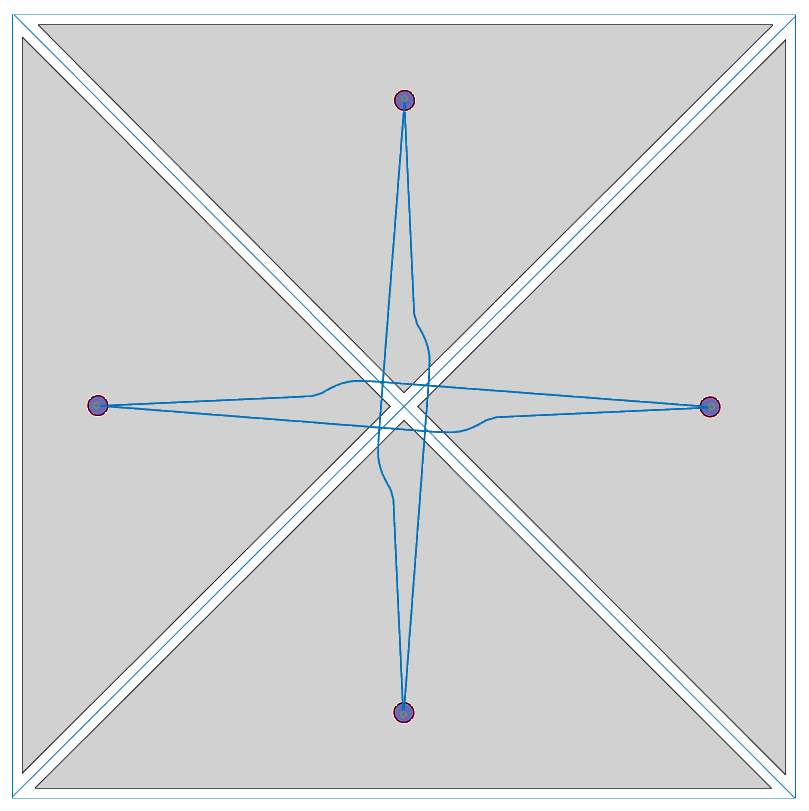}\hfill
  \includegraphics[width=.32\columnwidth]{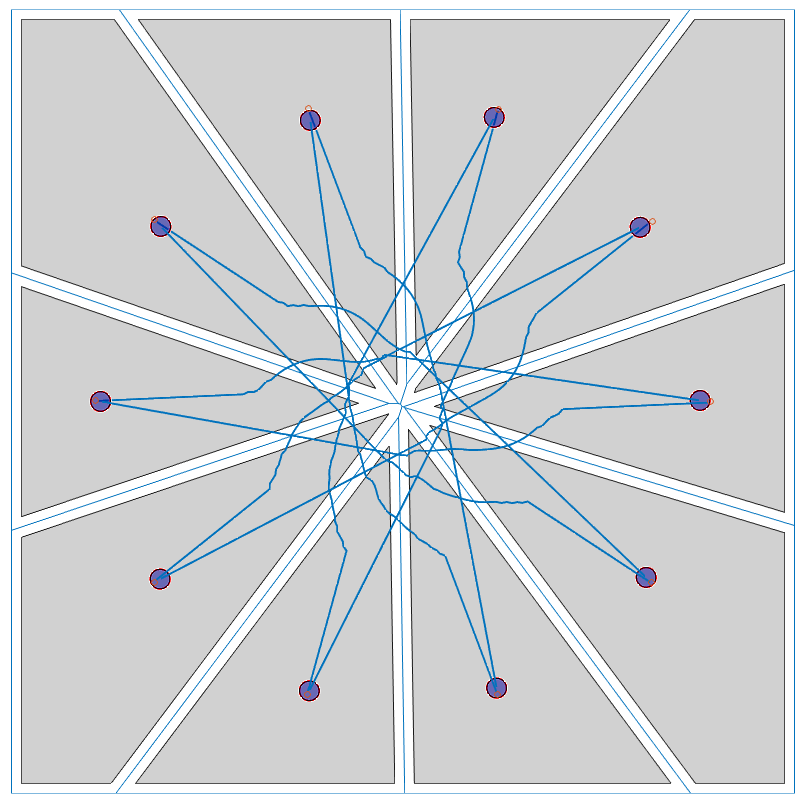}\hfill
    \includegraphics[width=.32\columnwidth]{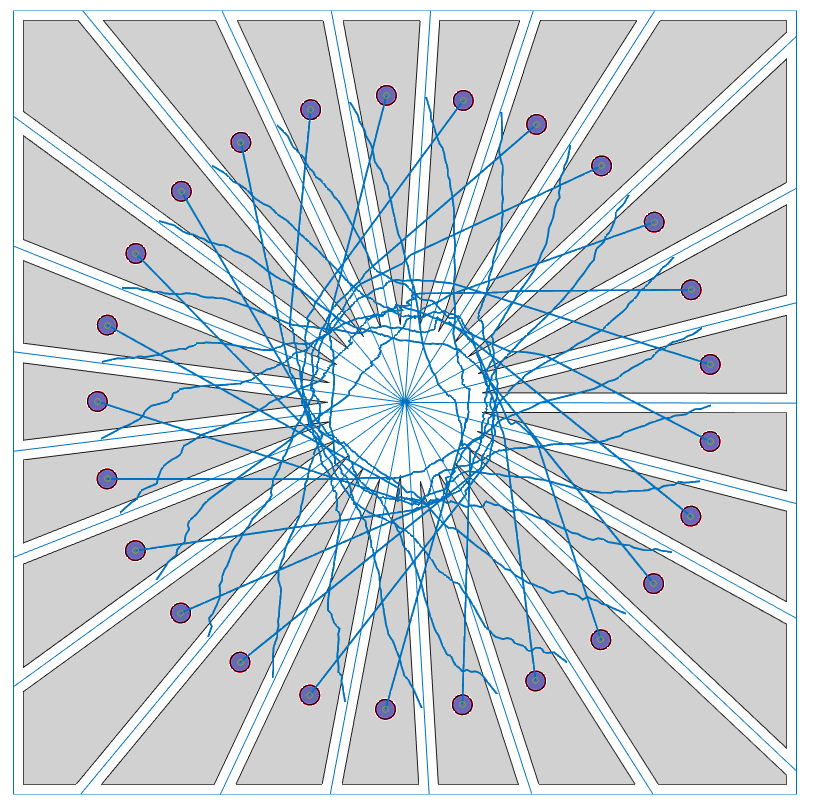}
  \caption{Collision-free trajectories for single integrator dynamics ($\dot{\mathbf{p}} = \mathbf{u}$). The sub images represent scenarios with 4 agents (left), 10 agent (center), and 25 agents (right). }
  \label{fig:SI}
\end{figure}

\begin{figure}[h]
  \includegraphics[width=.32\columnwidth]{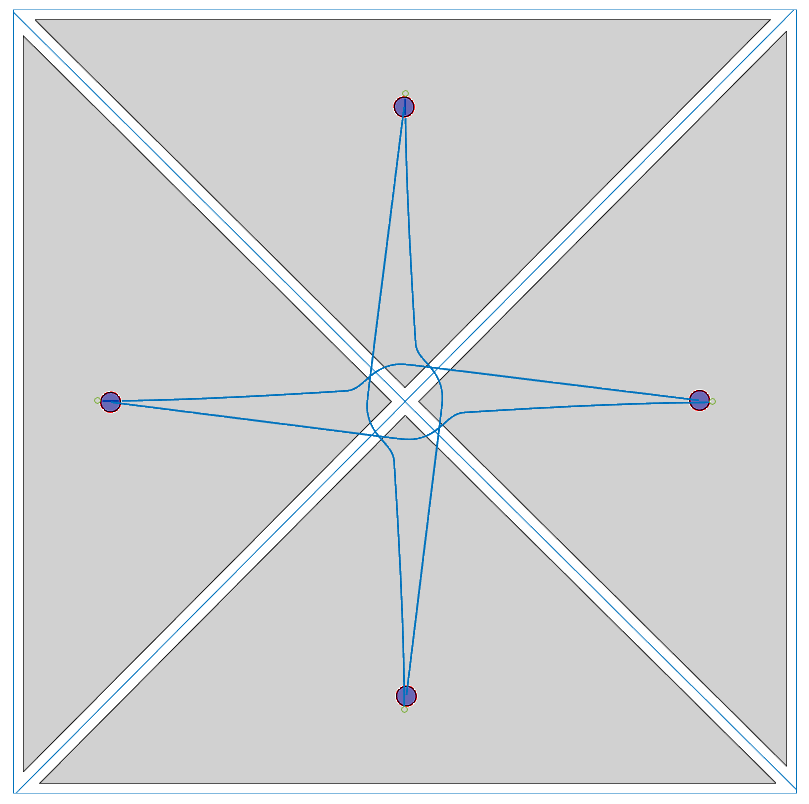}\hfill
  \includegraphics[width=.32\columnwidth]{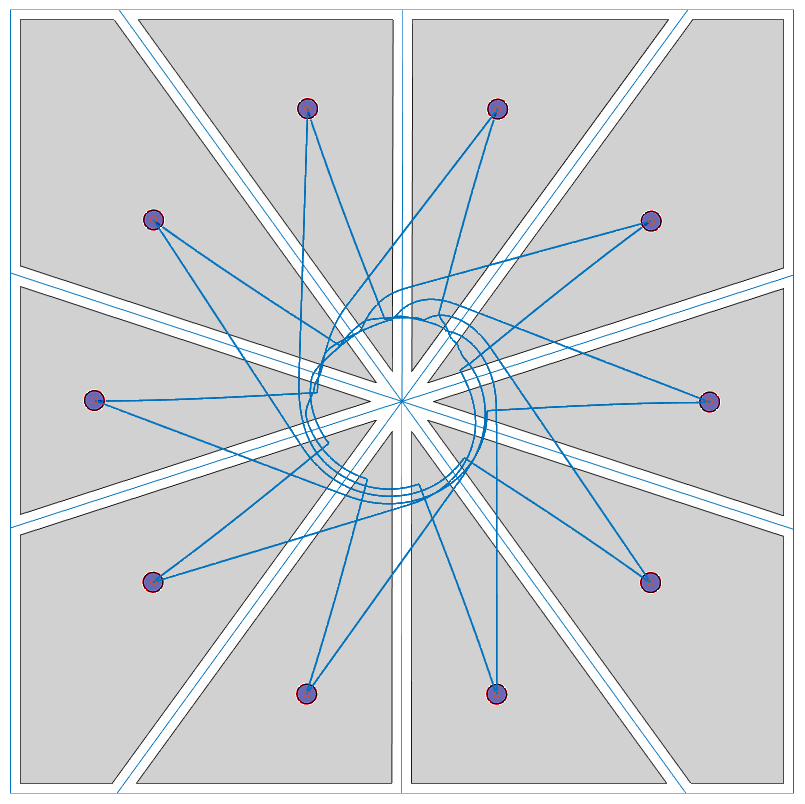}\hfill
    \includegraphics[width=.32\columnwidth]{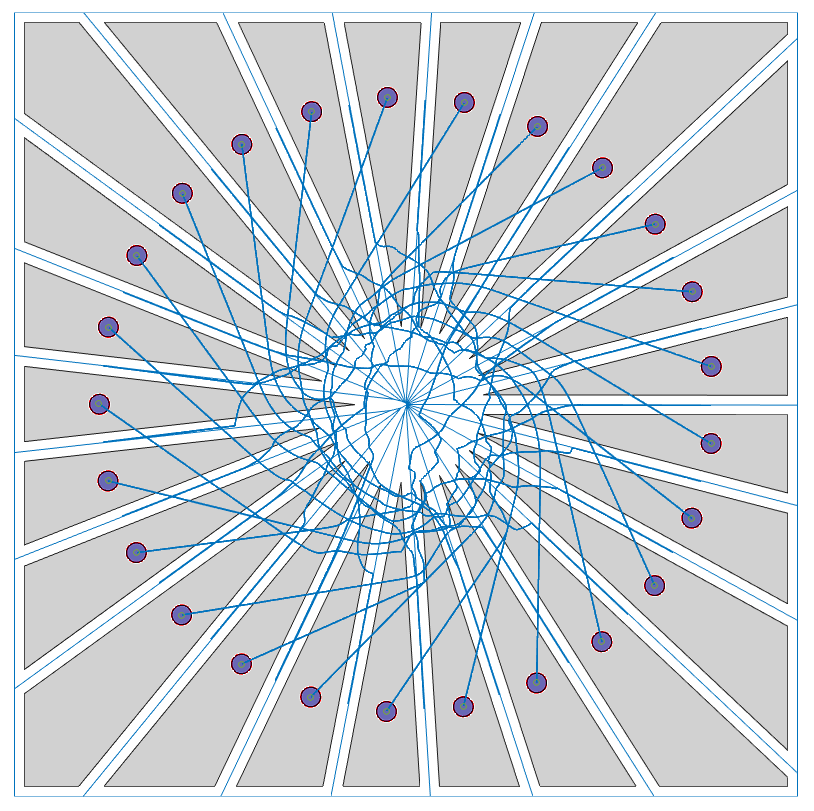}
  \caption{Collision-free trajectories for double integrator dynamics ($\ddot{\mathbf{p}} = \mathbf{u}$). The sub images represent scenarios with 4 agents (left), 10 agent (center), and 25 agents (right). ORCA cannot compute collision-free trajectories for the 25-agent scenario.}
  \label{fig:DI}
\end{figure}

We evaluate our method on simulation in a circular scenario, where each agent moves to its antipodal goal position. We vary the number of agents and increase the density in the scenario. Figures~(\ref{fig:SI}) and~(\ref{fig:DI}) illustrates the agent trajectories for single and double integrator dynamics, respectively.

\begin{figure}[ht]
\centering
  \includegraphics[width=.4\columnwidth]{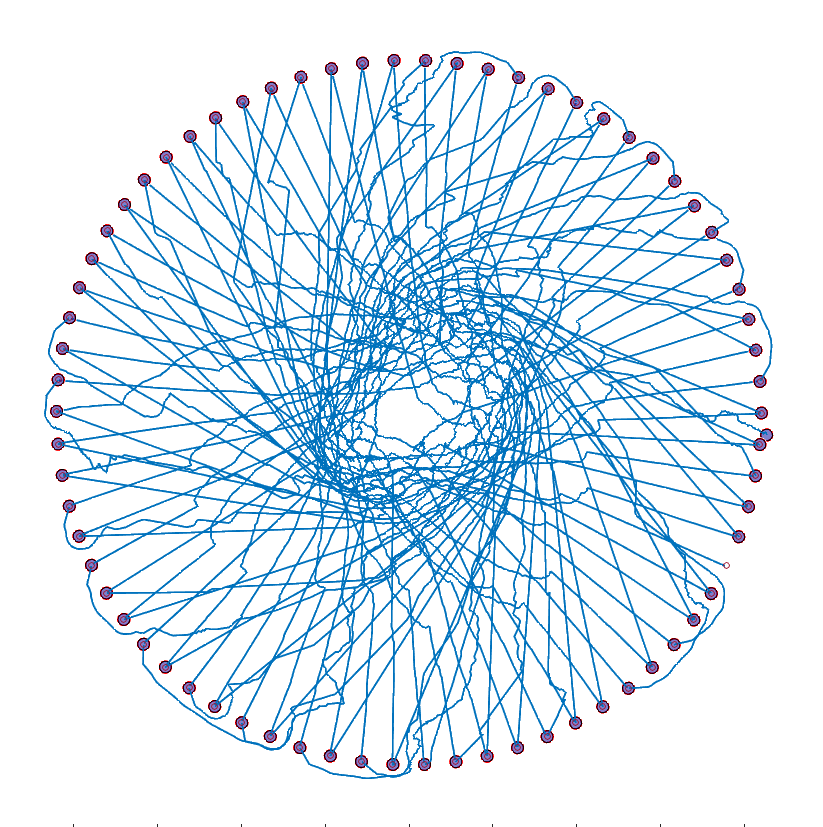} \quad
  \includegraphics[width=.4\columnwidth]{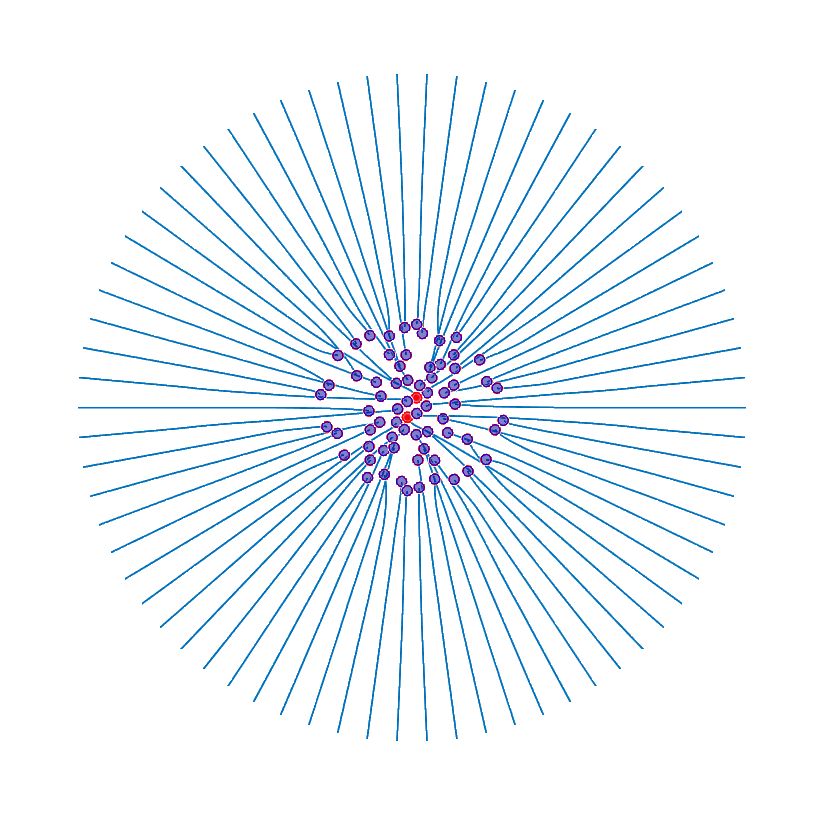}
  \caption{Trajectories for a scenario with $70$ agents. V-RVO (left) can generate collision-free trajectories for all the agents.  ORCA (right) results in collisions, as shown with red agents.
  }
  \label{fig:70}
\end{figure}

In addition, we compared V-RVO and ORCA on a scenario with 70 agents (Fig. \ref{fig:70}). While V-RVO can handle such complex scenario, ORCA results in collisions. This is due to the fact that ORCA constraints are not satisfied and it computes the closest velocity, which may not guarantee collision-free navigation.

\begin{figure}
\centering
  \includegraphics[width=.3\columnwidth, height=2.4cm, trim={3cm 8cm 3cm 8cm},clip]{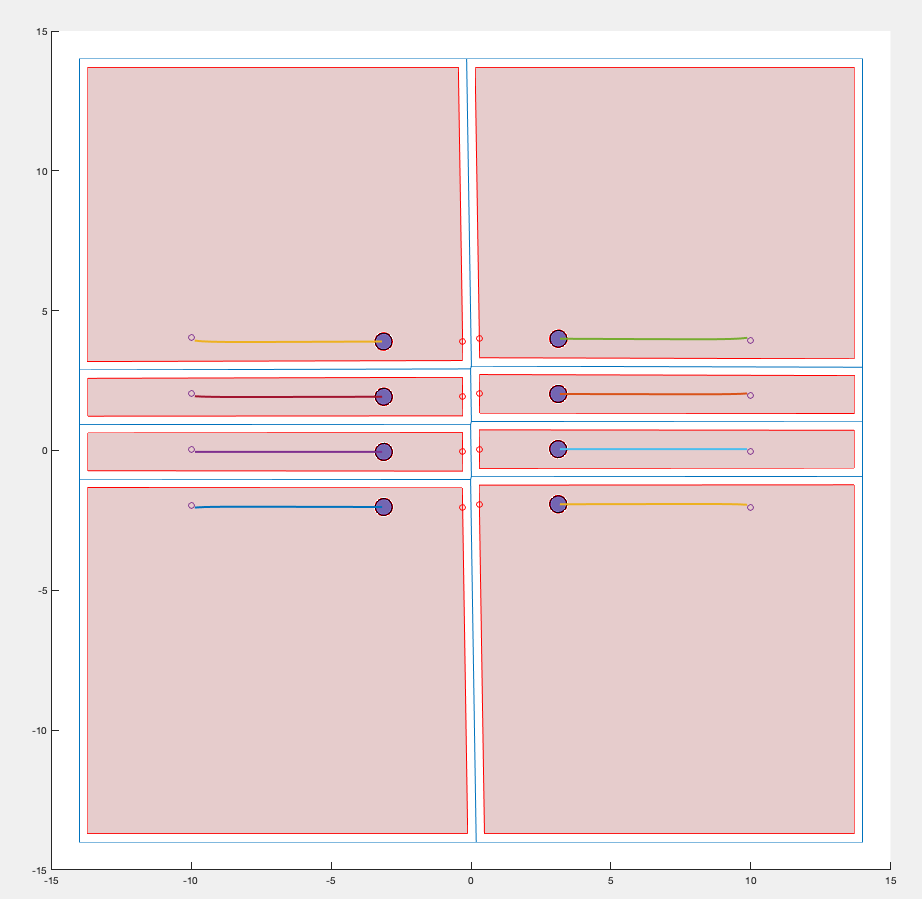} 
  \includegraphics[width=.3\columnwidth,  height=2.4cm, trim={3cm 8cm 3cm 8cm},clip]{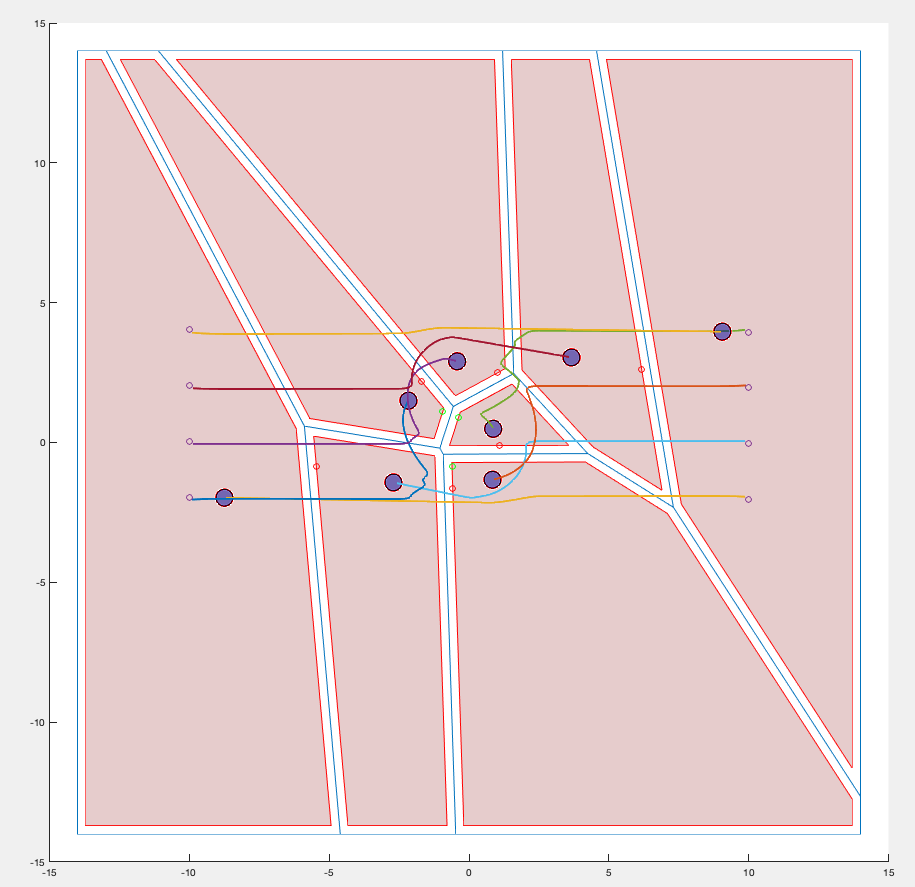}
    \includegraphics[width=.3\columnwidth,  height=2.4cm, trim={3cm 8cm 3cm 8cm},clip]{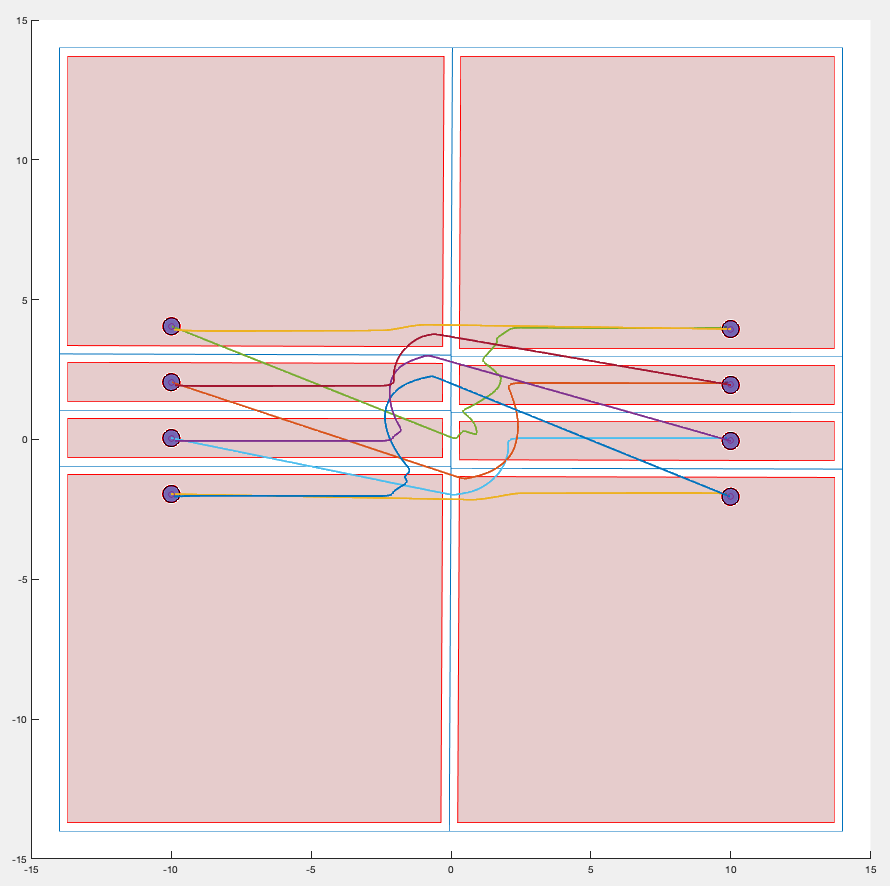}

\caption{We highlight the agent trajectories computed using V-RVO with 4 pairs. The pink regions denote the BVCs associated with each agent during each time step. The left, center, and right images show the BVCs and the trajectories in chronological order.}
\label{fig:approach}
\end{figure}

We also evaluated V-RVO in a scenarios with agents approaching with head on collisions. Figure~\ref{fig:approach} illustrates the collision-free trajectories.

\begin{figure}[t]
\centering
 \begin{subfigure}{.98\linewidth}
   \begin{subfigure}{.49\linewidth}
     \includegraphics[width=.99\columnwidth, trim={1cm 8cm 1cm 8cm},clip]{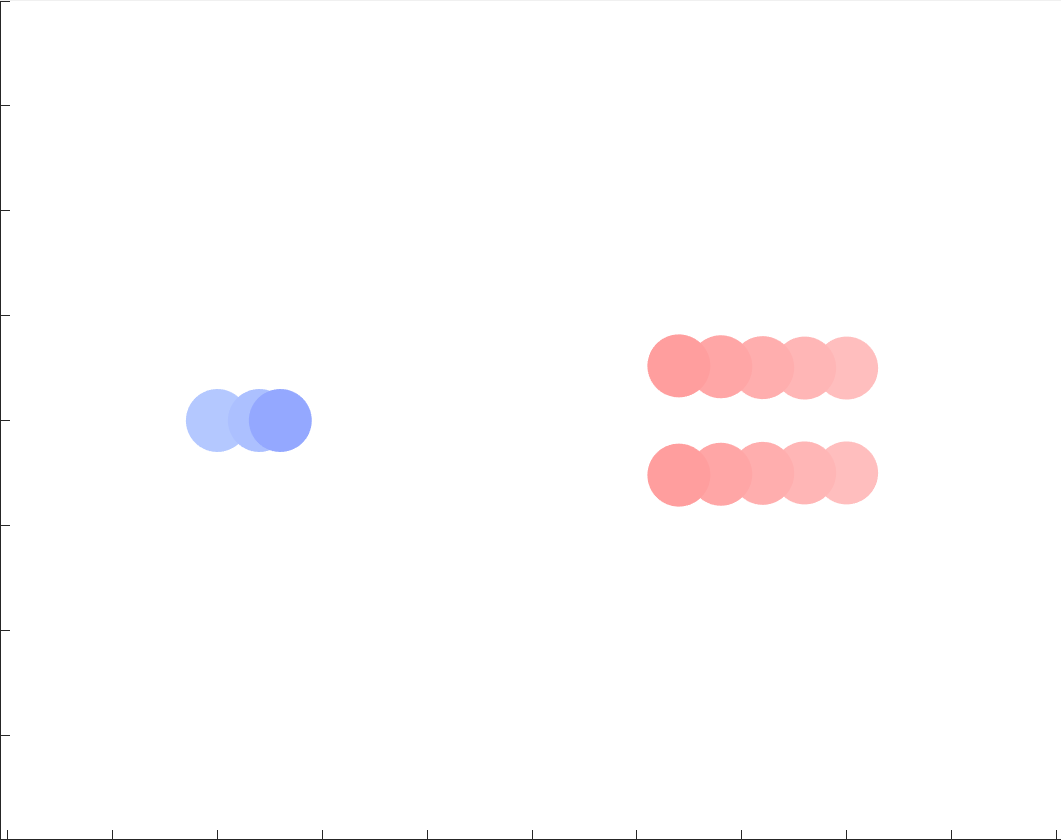} 
     \caption{t = 4 seconds}
   \end{subfigure}
   \begin{subfigure}{.49\linewidth}
     \includegraphics[width=.99\columnwidth, trim={1cm 8cm 1cm 8cm},clip]{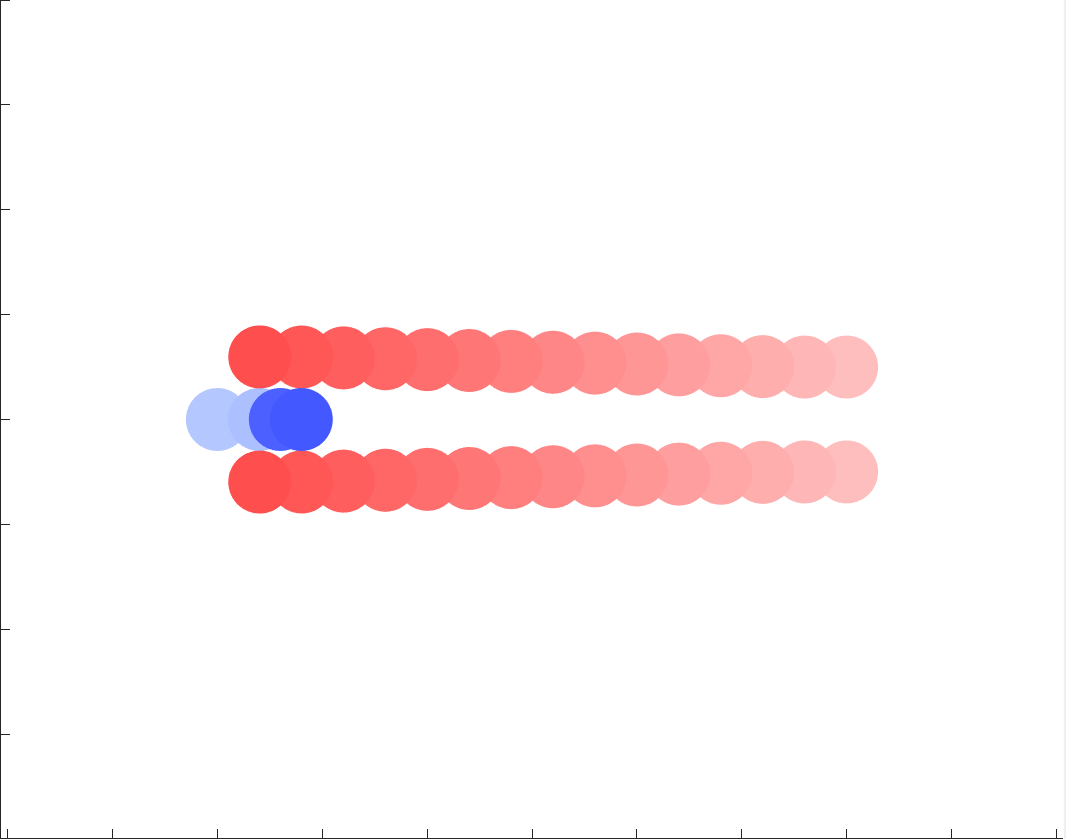}
     \caption{t = 8 seconds}
   \end{subfigure}
   \caption{ORCA}
   \label{fig:orca_2vs1}
 \end{subfigure}
 
 \begin{subfigure}{.98\linewidth}
   \begin{subfigure}{.49\linewidth}
     \includegraphics[width=.99\columnwidth, trim={1cm 8cm 1cm 8cm},clip]{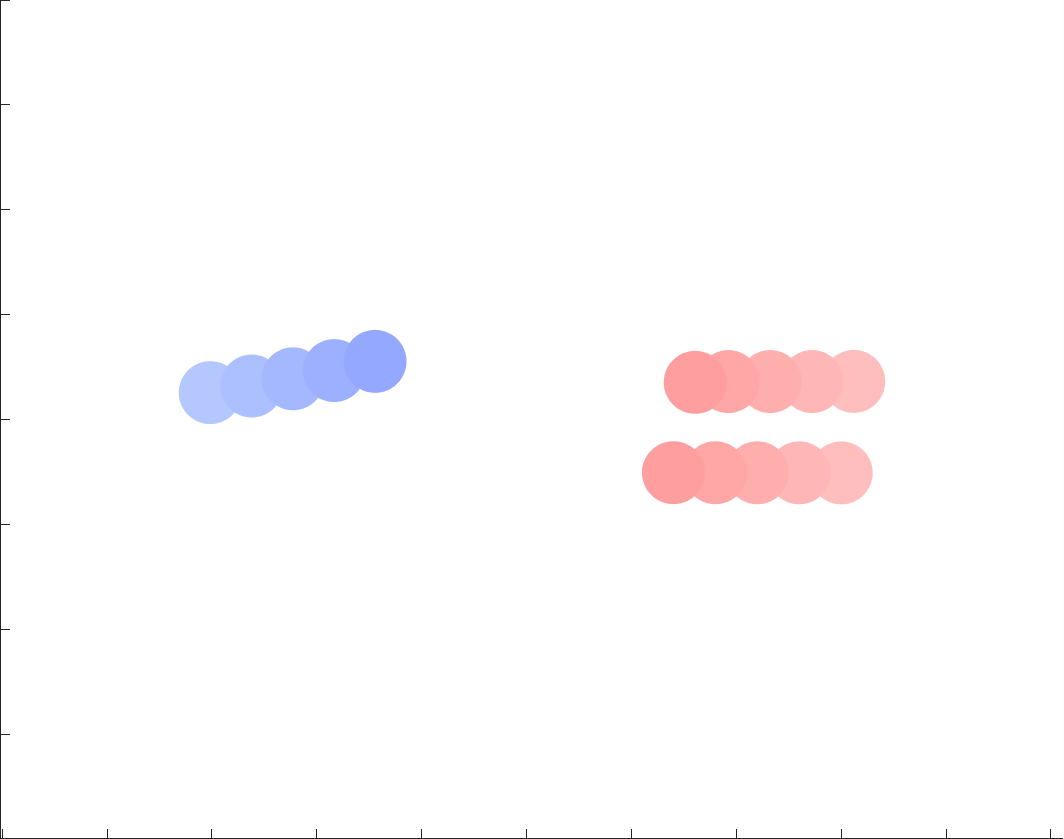} 
     \caption{t = 4 seconds}
   \end{subfigure}
   \begin{subfigure}{.49\linewidth}
     \includegraphics[width=.99\columnwidth, trim={1cm 8cm 1cm 8cm},clip]{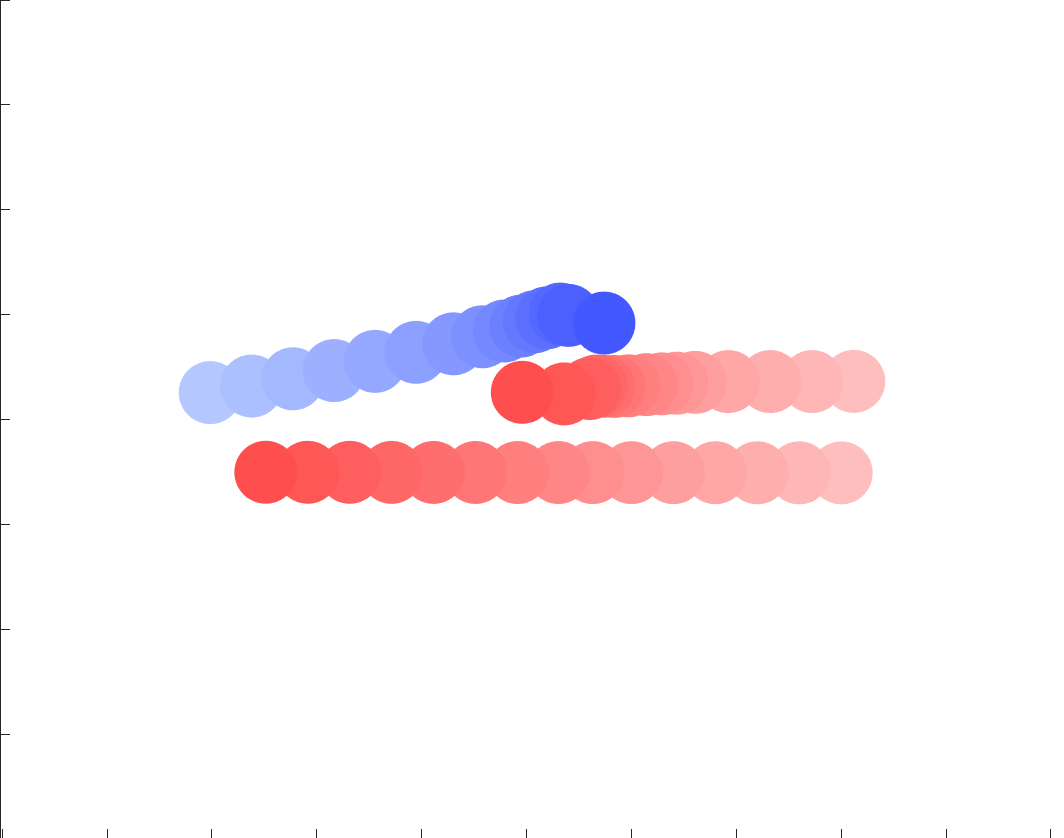} 
     \caption{t = 8 seconds}
   \end{subfigure}
  \caption{VRVO}
  \label{fig:vrvo2vs1}
 \end{subfigure}
    \caption{Image illustrates a three agent scenario where two {\em{red}} agents travel from right to left, while the {\em{blue}} agent travels from left to right. The trajectory of the agents are shown using colored disks where the agent's positions for recent time steps are denoted by darker color shades. While using ORCA, the {\em{blue}} agent is stays at the same position for several time steps due to ORCA's conservative nature. In contrast, while using VRVO the {\em{blue}} agent deviates to avoid a collision.}
\end{figure}

\subsection{Deadlock Resolution}\label{sec:deadlock}
\begin{figure}[h]
  \centering
  \includegraphics[width=.35\columnwidth,height=2.8cm,trim={5cm 5cm 5cm 5cm},clip]{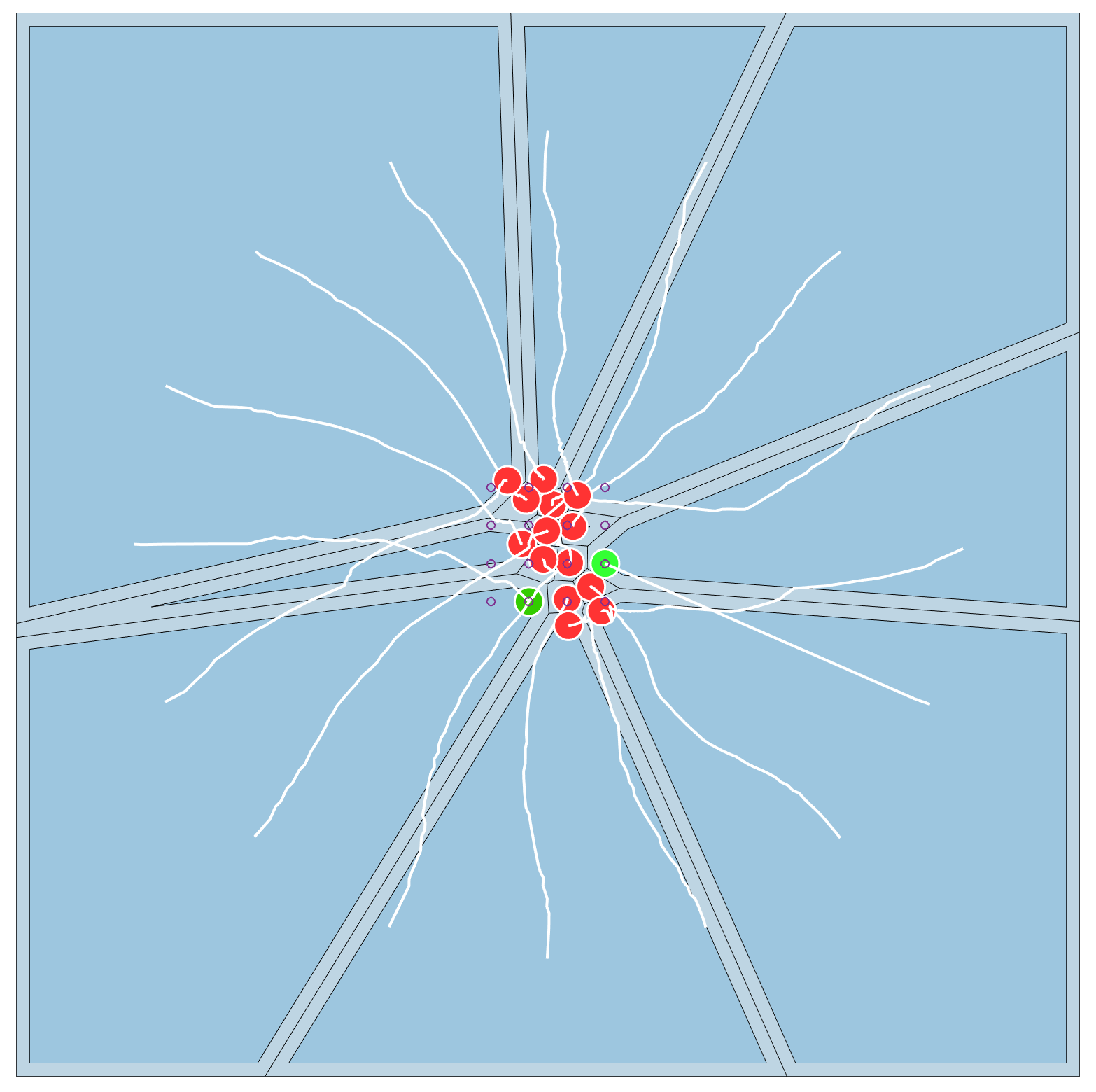}
  \includegraphics[width=.35\columnwidth,height=2.8cm,trim={5cm 5cm 5cm 5cm},clip]{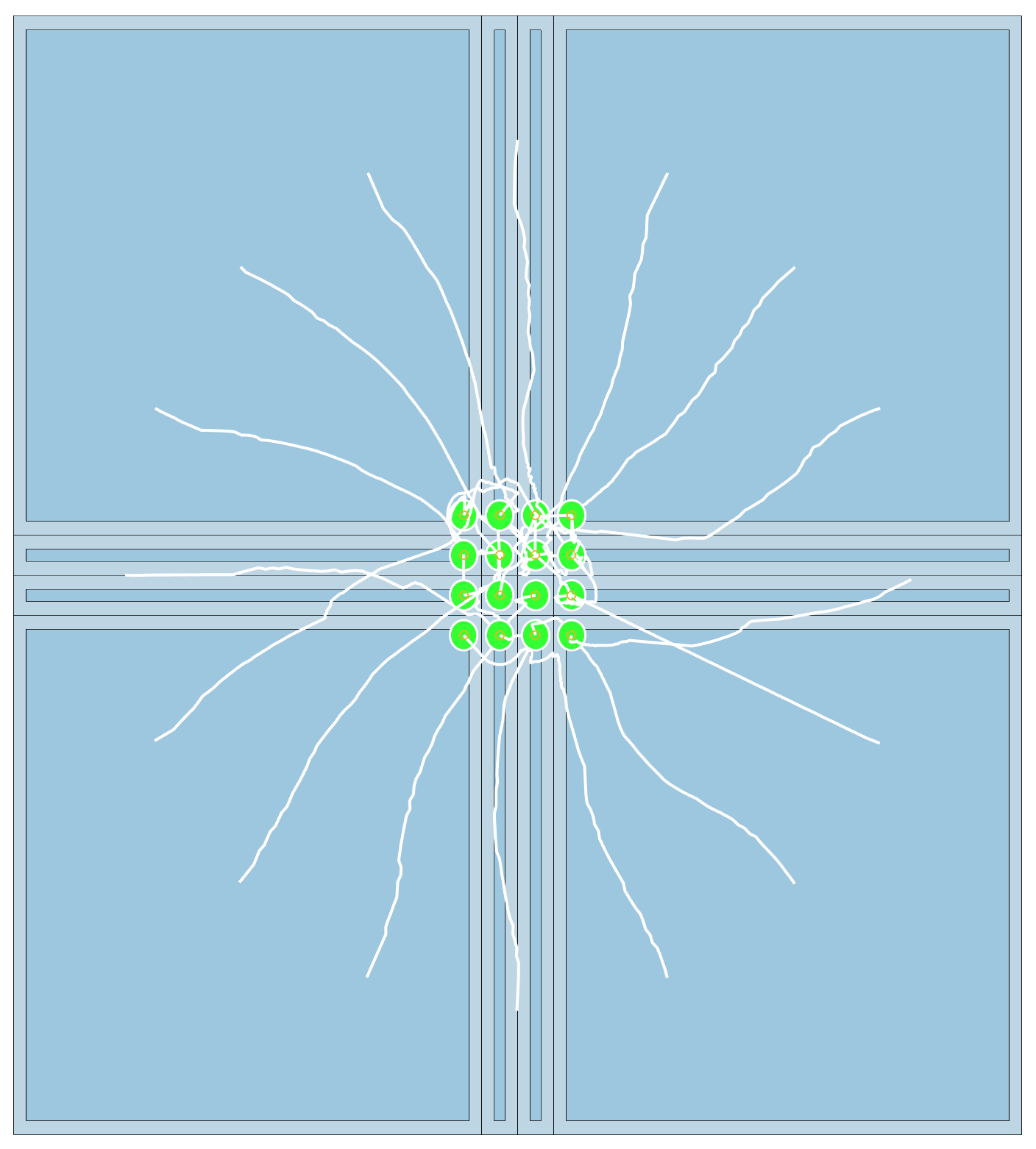}
  \caption{We highlight the performance of V-RVO algorithm without deadlock avoidance (left) and with deadlock avoidance (right).  The adjacent goal locations in the grid are $1.25$m apart. The agents in {\em green} have reached there goal, while the agent in {\em red} are deadlocked, as shown with their white trajectories.
  }
  \label{fig:deadlockgrid}
\end{figure}

We evaluate the performance of our deadlock resolution method on the formation benchmark. There are $16$ agents and they are initially placed on the circle perimeter equally spaced. The final positions of the agents are arranged in a $4 \times 4$ grid formation (see Fig.~\ref{fig:deadlockgrid}).  Our deadlock resolution method works well and the agents can reach the goal position in Fig.~\ref{fig:deadlockgrid} (right). 

\subsection{Scalability}
We compare the time required to compute the control inputs for the agent in scenarios with a team size ranging from $5$ to $70$ agents. Each agent considers only its neighbors within its sensing region for the computation. The average time per agent is in the range $1-2$ms and the overall performance is almost linear in the number of agents 

\begin{figure}
    \centering
    \includegraphics[width=0.7\columnwidth]{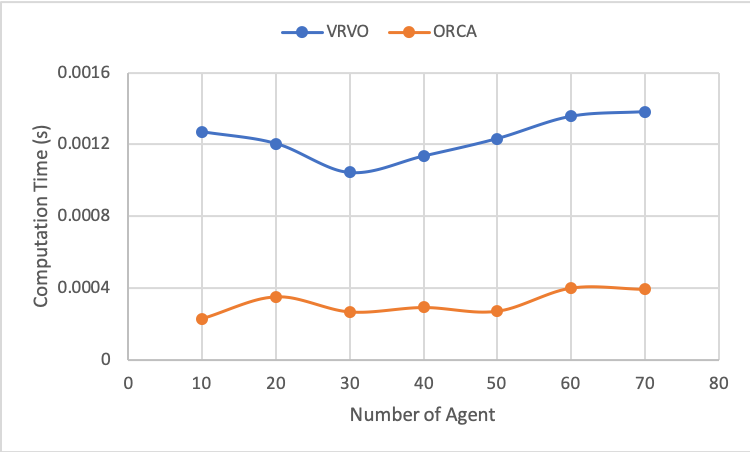}
    \caption{Average time per agent taken by V-RVO vs ORCA, as we vary the number of agents between $5-70$. V-RVO has additional overhead of computing Voronoi regions and baBVC and is about $3X$ slower than ORCA.
    }
    \label{fig:my_label}
\end{figure}

\section{Conclusion and Future Work}

In this paper, we presented a  new decentralized method based on Voronoi diagrams and RVO that computes a collision-free path for agents operating in a shared workspace with other agents, static and dynamic obstacles. Our V-RVO algorithm provides passive collision avoidance guarantees and we demonstrates its performance on agents with first and second-order dynamics. We also presented a deadlock resolution strategy and demonstrates its benefits over prior velocity-obstacle method.

Our method has some limitations. Our approach can be conservative and not complete in terms of always finding a collision-free trajectory (if that exists). Furthermore, our method does not guarantee deadlock resolution, similar to  prior decentralized methods. Our Voronoi and baBVC implementations are not optimized and can be accelerated. As a part of our future work, we plan to explore extending our deadlock resolutions and further improve the performance. We would also like to handle agents with non-holonomic constraints or imprecise state information (e.g., due to sensor errors).

   
\bibliographystyle{IEEEtran}
\bibliography{IEEEabrv,IEEEbib}

\begin{thebibliography}{10}
\providecommand{\url}[1]{#1}
\csname url@rmstyle\endcsname
\providecommand{\newblock}{\relax}
\providecommand{\bibinfo}[2]{#2}
\providecommand\BIBentrySTDinterwordspacing{\spaceskip=0pt\relax}
\providecommand\BIBentryALTinterwordstretchfactor{4}
\providecommand\BIBentryALTinterwordspacing{\spaceskip=\fontdimen2\font plus
\BIBentryALTinterwordstretchfactor\fontdimen3\font minus
  \fontdimen4\font\relax}
\providecommand\BIBforeignlanguage[2]{{%
\expandafter\ifx\csname l@#1\endcsname\relax
\typeout{** WARNING: IEEEtran.bst: No hyphenation pattern has been}%
\typeout{** loaded for the language `#1'. Using the pattern for}%
\typeout{** the default language instead.}%
\else
\language=\csname l@#1\endcsname
\fi
#2}}

\bibitem{lifelong}
J.~Li, A.~Tinka, S.~Kiesel, J.~W. Durham, T.~Kumar, and S.~Koenig, ``Lifelong
  multi-agent path finding in large-scale warehouses,'' \emph{arXiv preprint
  arXiv:2005.07371}, 2020.

\bibitem{tang2018complete}
S.~Tang and V.~Kumar, ``A complete algorithm for generating safe trajectories
  for multi-robot teams,'' in \emph{Robotics Research}.\hskip 1em plus 0.5em
  minus 0.4em\relax Springer, 2018, pp. 599--616.

\bibitem{usc}
W.~{Hönig}, J.~A. {Preiss}, T.~K.~S. {Kumar}, G.~S. {Sukhatme}, and
  N.~{Ayanian}, ``Trajectory planning for quadrotor swarms,'' \emph{IEEE
  Transactions on Robotics}, vol.~34, no.~4, pp. 856--869, 2018.

\bibitem{dandrea}
M.~{Hamer}, L.~{Widmer}, and R.~{D’andrea}, ``Fast generation of
  collision-free trajectories for robot swarms using gpu acceleration,''
  \emph{IEEE Access}, vol.~7, pp. 6679--6690, 2019.

\bibitem{solovey2016finding}
K.~Solovey, O.~Salzman, and D.~Halperin, ``Finding a needle in an exponential
  haystack: Discrete rrt for exploration of implicit roadmaps in multi-robot
  motion planning,'' \emph{The International Journal of Robotics Research},
  vol.~35, no.~5, pp. 501--513, 2016.

\bibitem{goldenberg2014enhanced}
M.~Goldenberg, A.~Felner, R.~Stern, G.~Sharon, N.~Sturtevant, R.~C. Holte, and
  J.~Schaeffer, ``Enhanced partial expansion a,'' \emph{Journal of Artificial
  Intelligence Research}, vol.~50, pp. 141--187, 2014.

\bibitem{RVO}
J.~{van den Berg}, {Ming Lin}, and D.~{Manocha}, ``Reciprocal velocity
  obstacles for real-time multi-agent navigation,'' in \emph{2008 IEEE
  International Conference on Robotics and Automation}, 2008, pp. 1928--1935.

\bibitem{ORCA}
J.~Van Den~Berg, S.~J. Guy, M.~Lin, and D.~Manocha, ``Reciprocal n-body
  collision avoidance,'' in \emph{Robotics research}.\hskip 1em plus 0.5em
  minus 0.4em\relax Springer, 2011, pp. 3--19.

\bibitem{zhou2017fast}
D.~Zhou, Z.~Wang, S.~Bandyopadhyay, and M.~Schwager, ``Fast, on-line collision
  avoidance for dynamic vehicles using buffered voronoi cells,'' \emph{IEEE
  Robotics and Automation Letters}, vol.~2, no.~2, pp. 1047--1054, 2017.

\bibitem{davis2019nh}
B.~Davis, I.~Karamouzas, and S.~J. Guy, ``Nh-ttc: A gradient-based framework
  for generalized anticipatory collision avoidance,'' \emph{arXiv preprint
  arXiv:1907.05945}, 2019.

\bibitem{VO}
P.~Fiorini and Z.~Shiller, ``Motion planning in dynamic environments using
  velocity obstacles,'' \emph{The International Journal of Robotics Research},
  vol.~17, no.~7, pp. 760--772, 1998.

\bibitem{best2016real}
A.~Best, S.~Narang, and D.~Manocha, ``Real-time reciprocal collision avoidance
  with elliptical agents,'' in \emph{2016 IEEE International Conference on
  Robotics and Automation (ICRA)}.\hskip 1em plus 0.5em minus 0.4em\relax IEEE,
  2016, pp. 298--305.

\bibitem{AVO}
J.~Van Den~Berg, J.~Snape, S.~J. Guy, and D.~Manocha, ``Reciprocal collision
  avoidance with acceleration-velocity obstacles,'' in \emph{2011 IEEE
  International Conference on Robotics and Automation}.\hskip 1em plus 0.5em
  minus 0.4em\relax IEEE, 2011, pp. 3475--3482.

\bibitem{LQG}
J.~Van Den~Berg, D.~Wilkie, S.~J. Guy, M.~Niethammer, and D.~Manocha,
  ``Lqg-obstacles: Feedback control with collision avoidance for mobile robots
  with motion and sensing uncertainty,'' in \emph{2012 IEEE International
  Conference on Robotics and Automation}.\hskip 1em plus 0.5em minus
  0.4em\relax IEEE, 2012, pp. 346--353.

\bibitem{LQR}
D.~Bareiss and J.~Van~den Berg, ``Reciprocal collision avoidance for robots
  with linear dynamics using lqr-obstacles,'' in \emph{2013 IEEE International
  Conference on Robotics and Automation}.\hskip 1em plus 0.5em minus
  0.4em\relax IEEE, 2013, pp. 3847--3853.

\bibitem{NH-ORCA}
\BIBentryALTinterwordspacing
J.~Alonso-Mora, A.~Breitenmoser, M.~Rufli, P.~Beardsley, and R.~Siegwart,
  \emph{Optimal Reciprocal Collision Avoidance for Multiple Non-Holonomic
  Robots}.\hskip 1em plus 0.5em minus 0.4em\relax Berlin, Heidelberg: Springer
  Berlin Heidelberg, 2013, pp. 203--216. [Online]. Available:
  \url{https://doi.org/10.1007/978-3-642-32723-0\_15}
\BIBentrySTDinterwordspacing

\bibitem{kim2015brvo}
S.~Kim, S.~J. Guy, W.~Liu, D.~Wilkie, R.~W. Lau, M.~C. Lin, and D.~Manocha,
  ``Brvo: Predicting pedestrian trajectories using velocity-space reasoning,''
  \emph{The International Journal of Robotics Research}, vol.~34, no.~2, pp.
  201--217, 2015.

\bibitem{he2017efficient}
L.~He, J.~Pan, and D.~Manocha, ``Efficient multi-agent global navigation using
  interpolating bridges,'' in \emph{2017 IEEE International Conference on
  Robotics and Automation (ICRA)}.\hskip 1em plus 0.5em minus 0.4em\relax IEEE,
  2017, pp. 4391--4398.

\bibitem{zhuBVC}
H.~Zhu and J.~Alonso-Mora, ``B-uavc: Buffered uncertainty-aware voronoi cells
  for probabilistic multi-robot collision avoidance,'' in \emph{2019
  International Symposium on Multi-Robot and Multi-Agent Systems (MRS)}.\hskip
  1em plus 0.5em minus 0.4em\relax IEEE, 2019, pp. 162--168.

\bibitem{desaraju2011decentralized}
V.~R. Desaraju and J.~P. How, ``Decentralized path planning for multi-agent
  teams in complex environments using rapidly-exploring random trees,'' in
  \emph{2011 IEEE International Conference on Robotics and Automation}.\hskip
  1em plus 0.5em minus 0.4em\relax IEEE, 2011, pp. 4956--4961.

\bibitem{forootaninia2017uncertainty}
Z.~Forootaninia, I.~Karamouzas, and R.~Narain, ``Uncertainty models for
  ttc-based collision-avoidance,'' in \emph{Robotics: Science and Systems},
  vol.~7, 2017.

\bibitem{ICS}
T.~{Fraichard} and H.~{Asama}, ``Inevitable collision states. a step towards
  safer robots?'' in \emph{Proceedings 2003 IEEE/RSJ International Conference
  on Intelligent Robots and Systems (IROS 2003) (Cat. No.03CH37453)}, vol.~1,
  2003, pp. 388--393 vol.1.

\bibitem{ICSfov}
S.~{Bouraine}, T.~{Fraichard}, and H.~{Salhi}, ``Provably safe navigation for
  mobile robots with limited field-of-views in unknown dynamic environments,''
  in \emph{2012 IEEE International Conference on Robotics and Automation},
  2012, pp. 174--179.

\bibitem{macek2009}
K.~Macek, D.~A.~V. Govea, T.~Fraichard, and R.~Siegwart, ``Towards safe vehicle
  navigation in dynamic urban scenarios,'' 2009.

\end{thebibliography}


\end{document}